\documentclass{article}

\usepackage{arxiv}
\usepackage[utf8]{inputenc} 
\usepackage[T1]{fontenc}    

\usepackage{url}            
\usepackage{booktabs}       
\usepackage{amsfonts}       
\usepackage{nicefrac}       
\usepackage{microtype}      
\usepackage{lipsum}		
\usepackage{graphicx}
\usepackage{doi}

\title{IvyGPT: InteractiVe Chinese pathwaY language model in medical domain}

\author{
    {Rongsheng Wang} \\
	Faculty of Applied Sciences\\
	Macao Polytechnic University\\
	Macao, China \\
	\texttt{p2213046@mpu.edu.mo} \\
    \And
	{Yaofei Duan} \\
	Faculty of Applied Sciences\\
	Macao Polytechnic University\\
	Macao, China \\
	\texttt{p2213964@mpu.edu.mo} \\
    \And
	{ChanTong Lam} \\
	Faculty of Applied Sciences\\
	Macao Polytechnic University\\
	Macao, China \\
	\texttt{ctlam@mpu.edu.mo} \\
    \And
	{Jiexi Chen} \\
	Faculty of Applied Sciences\\
	Macao Polytechnic University\\
	Macao, China \\
	\texttt{jaychan0302@gmail.com} \\
    \And
	{Jiangsheng Xu} \\
	Opera inc\\
	China \\
	\texttt{blueshell@whu.edu.cn} \\
	\And
	{Haoming Chen} \\
	Faculty of Applied Sciences\\
	Macao Polytechnic University\\
	Macao, China \\
	\texttt{p2213065@mpu.edu.mo} \\
    \And
	{Xiaohong Liu} \\
	John Hopcroft Center\\
    Shanghai Jiao Tong University \\
	Shanghai, China \\
	\texttt{xiaohongliu@sjtu.edu.cn} \\
    \And
	{Patrick Cheong-Iao Pang} \\
	Faculty of Applied Sciences\\
	Macao Polytechnic University\\
	Macao, China \\
	\texttt{mail@patrickpang.net} \\
    \And
	{Tao Tan*} \\
	Faculty of Applied Sciences\\
	Macao Polytechnic University\\
	Macao, China \\
	\texttt{taotanjs@gmail.com} \\
}

\renewcommand{\shorttitle}{\textit{arXiv} Template}

\hypersetup{
pdftitle={A template for the arxiv style},
pdfsubject={q-bio.NC, q-bio.QM},
pdfauthor={David S.~Hippocampus, Elias D.~Striatum},
pdfkeywords={First keyword, Second keyword, More},
}

\begin{document}
\maketitle

\begin{abstract}
General large language models (LLMs) such as ChatGPT have shown remarkable success. However, such LLMs have not been widely adopted for medical purposes, due to poor accuracy and inability to provide medical advice. We propose IvyGPT, an LLM based on LLaMA that is trained and fine-tuned with high-quality medical question-answer (QA) instances and Reinforcement Learning from Human Feedback (RLHF). After supervised fine-tuning, IvyGPT has good multi-turn conversation capabilities, but it cannot perform like a doctor in other aspects, such as comprehensive diagnosis. Through RLHF, IvyGPT can output richer diagnosis and treatment answers that are closer to human. In the training, we used QLoRA to train 33 billion parameters on a small number of NVIDIA A100 (80GB) GPUs. Experimental results show that IvyGPT has outperformed other medical GPT models. 


\end{abstract}

\section{Introduction}
Large language models (LLMs) are an important research direction in the field of artificial intelligence and have seen rapid growth after the introduction of ChatGPT in 2022. These large language models are trained based on large amounts of text data and are capable of generating human-like text, answering questions, and performing other natural language processing (NLP)-related tasks with high accuracy, showing excellent performance in a wide range of tasks. A key development in this area is the use of the Transformer architecture~\cite{1-2} and the underlying attention mechanism~\cite{1-3}, which greatly improves the ability of language models to handle long-range dependencies in natural language text. Another important development is the use of pre-training, using LLMs that have been trained on large datasets and then fine-tuned for specific tasks, allowing the models to perform better in domain-specific Q\&A, text generation tasks, further enabling LLMs to provide more accurate support for specialized domains or specific industries.

LLMs such as ChatGPT and GPT-4 are general-purpose models that are not trained to be used specifically for medical problems and are not designed to solve clinical tasks. Therefore, the use of LLMs in the medical domain may lead to incorrect or misleading medical information, and such incorrect or incomplete information may have harmful consequences for patient care. In addition, due to the diversity of global medical approaches (e.g., Chinese and Western medicine), LLMs for use in the medical domain need to be adapted to local medical practices and treatment methods. Currently, common LLMs are typically trained in English, which limits their ability to understand and respond semantically to other languages and affects the ability of models to reason in populations that do not speak English as their first language.

LLMs for use in the medical field have many advantages. They can provide diagnostic advice, medication recommendations, interpretation of medical reports, health care recommendations, and more. By providing basic information about patients through online chat, LLMs can perform online medical services based on the information provided. Through NLP technology, LLMs automatically sift and generate information about disease diagnosis, treatment, etc. From massive medical data to automate the processing of patient questions and needs. And the processing speed of large language models is so fast that they can respond to patients' needs quickly, ensuring that patients can get medical advice on time and effectively reducing the problem of patient congestion in hospitals~\cite{X}.

Several LLMs for the medical domain have been proposed for the Chinese language, including HuaTuo~\cite{1-4}, ChatMed~\cite{1-5}, ZhenNong-TCM~\cite{1-6}, and MedicalGPT (zh) ~\cite{1-7}. However, these models have shortcomings such as small parameter sizes (like 6B or 7B) and the lack of high-quality training data conforming to real doctor-patient scenarios. These limitations restrict their generalization abilities in medicine. To address these issues, we propose IvyGPT for the medical settings in China and our contributions in this paper include:
\begin{enumerate}
    \item We propose a comprehensive training method for medical LLMs which comprises three parts: supervised training, reward model training, and reinforcement learning using the QLoRA method to train large models with 33B parameters on devices with low computing power.
    \item We contribute a high-quality dataset containing verified and realistic scenarios of doctor-patient conversations.
    \item We evaluate the IvyGPT against other LLMs in the medical domain.
\end{enumerate}

\section{Methodology}
\begin{figure}
\centering
\includegraphics[width=0.95\textwidth]{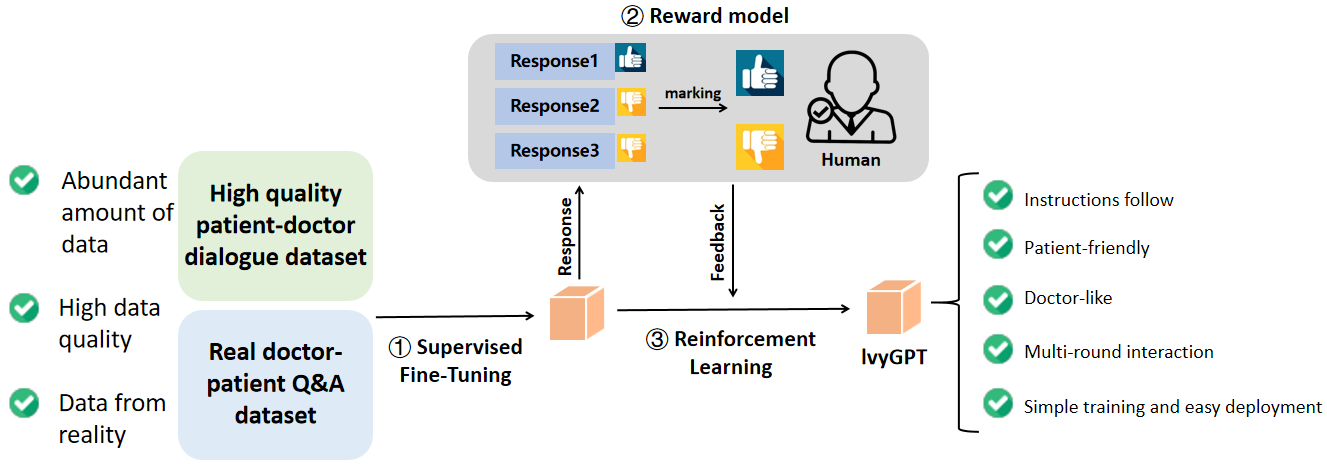}
\caption{Process strategy for IvyGPT training} 
\label{fig1}
\end{figure}

Our approach focuses on integrating high-quality data with human feedback to enhance the quality of responses in medical consultations. This is achieved through a two-stage training strategy: Supervised Fine-tuning (SFT) utilizing mixed data and Reinforcement Learning (RL) derived from human feedback, as illustrated in Fig.~\ref{fig1}.

\vspace{-0.5cm}

\subsection{Supervised Fine-Tuning}
LLMs are pre-trained on large-scale unlabeled data, so they have strong language understanding and generation capabilities. However, for medical question-and-answer tasks, models may require more guidance and tuning to achieve optimal performance. With supervised fine-tuning, the generalization ability of the model can be combined with the requirements of a specific task, thus improving task performance. However, fine-tuning large language models with full parameters is extremely resource-intensive. Hu et al.~\cite{3-4-2} proposed a method, LoRA, which would freeze the weights of pre-trained models and inject trainable rank decomposition matrices into each layer of the Transformer architecture, thus greatly reducing the number of trainable parameters required for downstream tasks. The QLoRA~\cite{3-4-1} method performs better on some fine-tuning tasks of large language models with larger parameters, and it will do a 4-bit quantization of the base model to ensure that the model can be fine-tuned for LLMs with very large parameters in a low memory footprint.

\subsection{Reinforcement Learning from Human Feedback}
\textbf{Reward Model} We train a reward model to align with the characteristics of doctors and LLMs. We use real instructions and conversations as training data, sampling multiple responses from our fine-tuned model. For multi-turn conversations, we provide the dialogue history to align our model’s response generation. These responses are then scored by human, considering informativeness, coherence, adherence to human preferences, and factual accuracy based on given real doctors’ diagnoses. The scoring LLM evaluates each response and assigns a score. We use this paired response data to train the reward model, using the fine-tuned model as its backbone for better generalization.

\textbf{Reinforcement Learning}
Throughout the training process, in order to ensure that the model's output is closer to human-preferred answers, we employ a reinforcement learning strategy to incentivize the model to produce better results. By defining a reward function, the system can learn to take appropriate actions during the dialogue process to maximize the expected reward. This allows the dialogue system to better understand user intent and provide accurate assistance. In the reinforcement learning process, we input the top k answers returned by the model for the same question into the reward model for scoring and obtain a score value. To prevent the model trained through reinforcement learning from deviating too far from the model in the SFT (Supervised Fine-Tuning) stage, we introduce constraint during training to ensure that the model achieves better results without veering off track. Without this constraint, the model may lean towards answers with extremely high scores, which may not necessarily be good answers.

\section{Experiments}

\subsection{Training Details}
Our model was implemented in PyTorch using the Accelerate, PEFT, and transformers packages, with LLaMA-33B~\cite{llama} as the base model. For training, we loaded the model onto 4 NVIDIA A100 80GB GPUs and performed fine-tuning. In the supervised fine-tuning process, we set the learning rate, batch size, and maximum context length to 5e-5, 16, and 1024, respectively. We trained all models for 3 epochs and saved the weights that performed the best on the validation set. During the reinforcement learning process, the model required larger memory usage. Therefore, we set the learning rate, batch size, and maximum context length to 5e-5, 8, and 256, respectively, and trained for 2 epochs. The model trained through reinforcement learning demonstrated improved understanding and generation of responses in various conversational scenarios, accurately following different instructions.

\subsection{Dataset}
First we used a dataset released from HuaTuoGPT~\cite{3-1}, and leveraged ChatGPT to verify grammatical errors and remove common sense errors. Second, we obtained a large number of real doctor-patient conversations from public websites and added them to the dataset. Finally, our dataset contains 307,038 sets of Q\&As. 

\begin{figure}[htbp]
\centering
{
    \begin{minipage}[b]{.47\linewidth}
        \centering
        \includegraphics[scale=0.5]{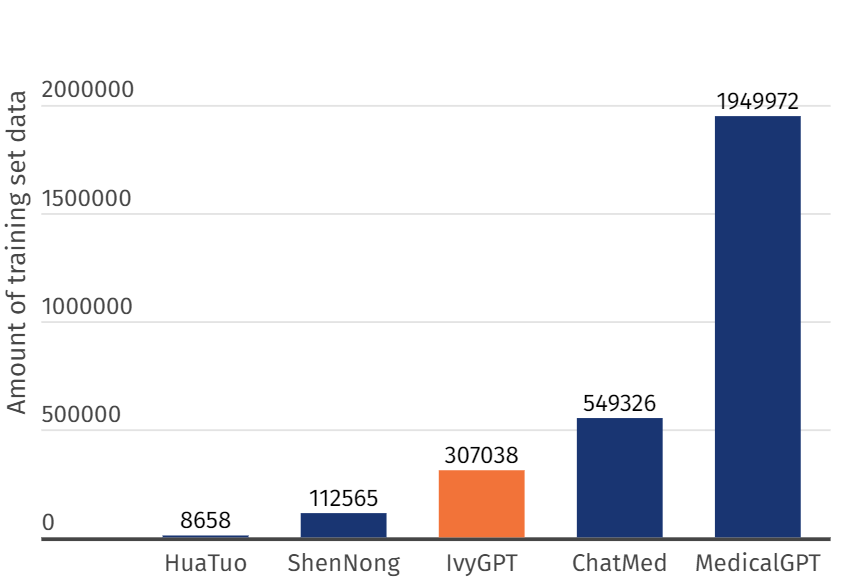}
    \end{minipage}
}
{
 	\begin{minipage}[b]{.47\linewidth}
        \centering
        \includegraphics[scale=0.47]{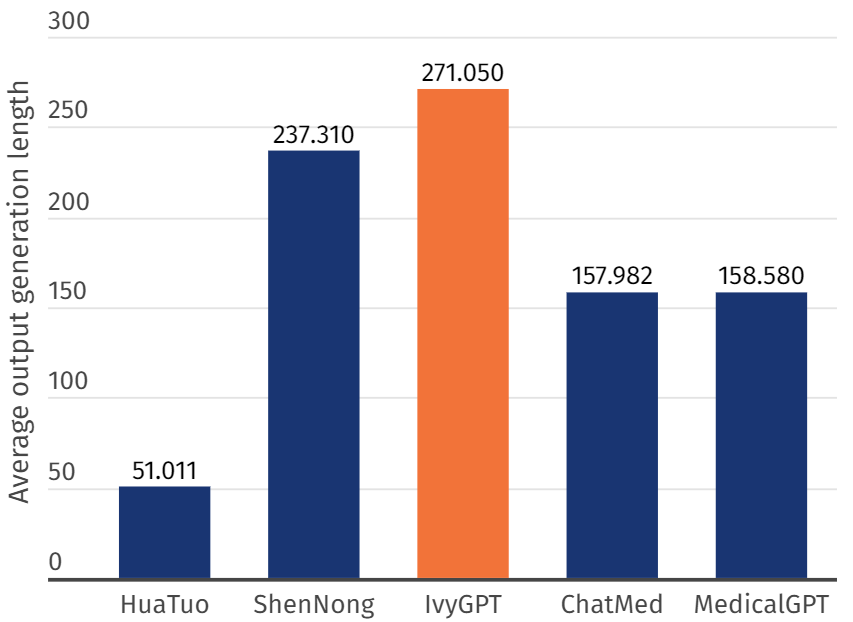}
    \end{minipage}
}
\caption{(a) The total amount of data in the training set, (b) The average number of words in the generated responses}
\label{f1}
\end{figure}

As shown in Fig.~\ref{f1} (a), we used a smaller amount of data to complete the training process of the model, which effectively reduces the training time and resource consumption. 
In Fig.~\ref{f1} (b), a comparison of the length of the model's response generated for the query is shown. IvyGPT generates an average word count of 271.05, which is higher than the current models in the Chinese medical domain. Usually, a higher word count of responses indicates that the model can cover richer information in the response.

\subsection{Evaluation Results}
We used the word2vec-based method of computing cosine similarity to evaluate semantic similarity of AI answers and human answers. We used the trained 64-dimensional word2vec model to embed answers. We compared IvyGPT with four proposed baseline models: HuaTuo, ShenNong, ChatMed, and MedicalGPT for analysis. We used 100 query pairs to evaluate the semantic similarity between ChatGPT and the Chinese medical domain model, respectively. The results are shown in Table~\ref{t1}, and our model obtains the highest semantic similarity. This implies that our model is better able to capture and express semantic relatedness, resulting in more accurate and consistent language generation.

\begin{table}[h]
\caption{Semantic similarity comparison with real doctor}
\begin{center}
\begin{tabular}{l|c}
\hline
\multicolumn{1}{c|}{Model} & Score \\ \hline
ShenNong~\cite{1-6}                   & 77.71                             \\ \hline
HuaTuo~\cite{1-4}                     & 71.20                             \\ \hline
ChatMed~\cite{1-5}                   & 84.51                             \\ \hline
MedicalGPT~\cite{1-7}                 & 83.73                             \\ \hline
ChatGPT            & 89.13                    \\ \hline
\textbf{IvyGPT (ours)}            & \textbf{93.58}                    \\ \hline
\end{tabular}
\label{t1}
\end{center}
\end{table}

\begin{figure}[htbp]
\centering
{
    \begin{minipage}[b]{.4\linewidth}
        \centering
        \includegraphics[scale=0.18]{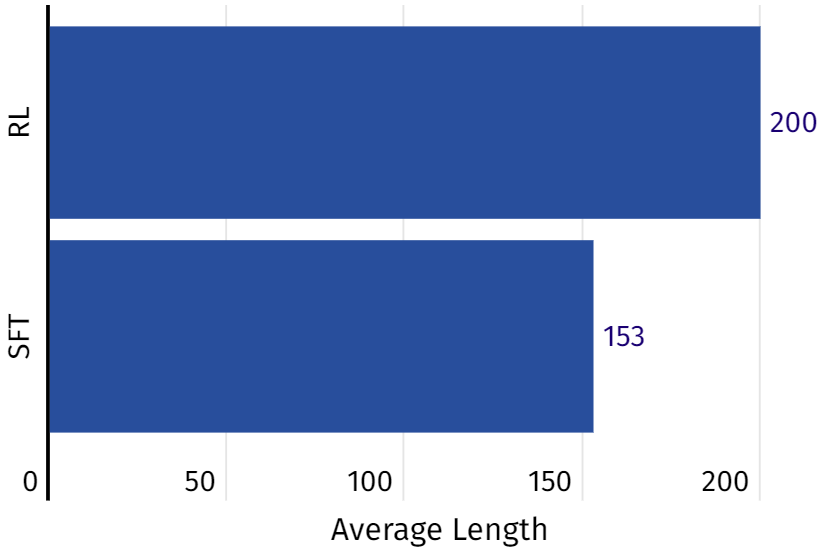}
        
    \end{minipage}
}
{
 	\begin{minipage}[b]{.4\linewidth}
        \centering
        \includegraphics[scale=0.3]{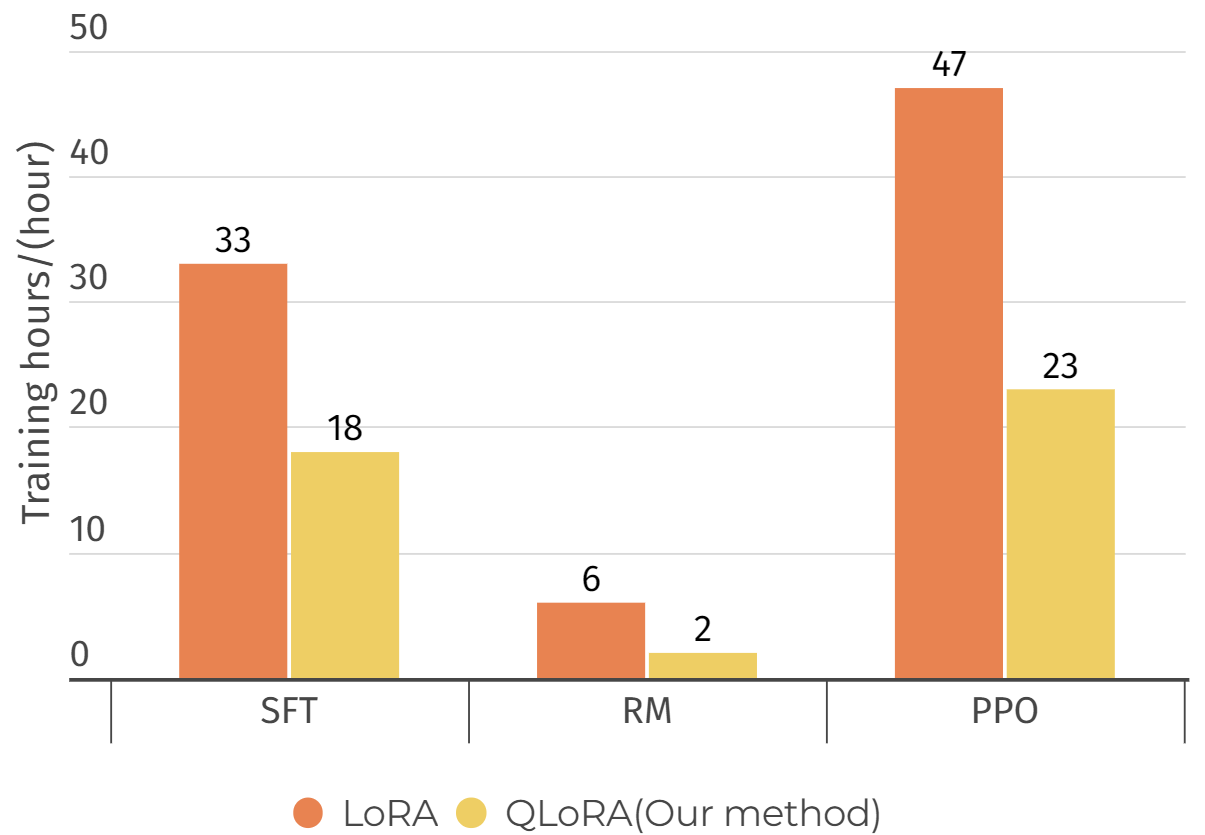}
    \end{minipage}
}
\caption{(a) The average word count of responses produced by the model after reinforcement learning. (b) The training duration under different fine-tuning methods}
\label{f2}
\end{figure}

We compare the output word count of the model with different training methods. As shown in the Fig.~\ref{f2} (a), the model with reinforcement learning has richer answers to users' questions. Fig.~\ref{f2} (b) shows that QLoRA has a shorter training time and is more efficient compared to LoRA.

\section{Discussion}

We proposes lvyGPT, a medical large-scale language model trained through supervised fine-tuning using high-quality medical question-and-answer instances and reinforcement learning based on human feedback. We demonstrate through experiments that lvyGPT generates responses with more reliable medical knowledge. However, there are still limitations in the application of lvyGPT in the medical field. Firstly, the difficulty in obtaining medical data limits the breadth and depth of the model's knowledge. Additionally, the model's ability to personalize and understand context is limited, leading to insufficient consideration of individual patient circumstances. Lastly, ethical and accountability issues need to be carefully addressed to ensure accuracy, reliability, and to avoid the spread of bias and misinformation. Therefore, while lvyGPT shows potential in the medical field, further improvements and validation are necessary.

There is still much work to be explored in the future, including:
\begin{enumerate}
    \item Combining large language models with other medical expert systems and knowledge bases can provide more comprehensive and accurate medical diagnostics and recommendations. By integrating multiple models and systems, their strengths can be fully utilized to improve overall performance and reliability.
    \item Explore various model compression techniques, such as pruning, quantization, distillation, etc., to reduce the size and computational requirements of models. This will make large language models easier to deploy and use, while lowering hardware requirements.
    \item Conducting thorough evaluations and trials in real-world medical settings to assess the model's performance, effectiveness, and impact on patient outcomes.
\end{enumerate}

\section{Conclusion}
We propose IvyGPT, a medical LLM trained through supervised fine-tuning using high-quality medical question-and-answer instances and RLHF. With QLoRA, IvyGPT is able to load and train a model with 33 billion parameters, even with limited computational resources. The results show that the responses generated by IvyGPT have a higher similarity to the Q\&A in real doctor-patient scenarios, indicating its potential in promoting self-service healthcare and supporting healthcare professionals.


\end{document}